%% file: main.tex
\pdfoutput=1

\documentclass[sigconf, screen]{acmart}

\usepackage{CJKutf8}
\usepackage{xcolor,soul}
\usepackage{colortbl}
\usepackage{graphicx} 
\usepackage{enumitem}
\usepackage{balance}
\usepackage{makecell}
\definecolor{mypink}{rgb}{.99,.90,.92}
\definecolor{magicmint}{rgb}{0.67, 0.94, 0.82}
\usepackage{caption}
\AtBeginDocument{%
  }

\renewcommand{\footnotetextcopyrightpermission}[1]{}  
\begin{document}
\begin{CJK}{UTF8}{gbsn}

\title{DEBATE: A Dataset for Disentangling Textual Ambiguity in Mandarin Through Speech}

\author{Haotian Guo}
\authornote{These authors contributed equally to this research.}
\email{haotianguo@hnu.edu.cn}
\affiliation{%
  \institution{Hunan University}
  \city{Changsha}
  \country{China}
}

\author{Jing Han}
\authornotemark[1]
\authornote{Jing Han, Weihao Gan, and Zixing Zhang are corresponding authors.}
\email{jh2298@cam.ac.uk}
\affiliation{%
  \institution{University of Cambridge}
  \city{Cambridge}
  \country{uk}
}

\author{Yongfeng Tu}
\authornotemark[1]
\email{tuyongfeng@mlslabs.com.cn}
\affiliation{%
  \institution{Malanshan Audio\&Video Laboratory}
  \city{Changsha}
  \country{China}
}

\author{Shihao Gao}
\email{shihaogao@hnu.edu.cn}
\affiliation{%
  \institution{Hunan University}
  \city{Changsha}
  \country{China}
}

\author{Shengfan Shen}
\email{shenshengfan@hnu.edu.cn}
\affiliation{%
  \institution{Hunan University}
  \city{Changsha}
  \country{China}
}

\author{Wulong Xiang}
\email{xiangwulong@hnu.edu.cn}
\affiliation{%
  \institution{Hunan University}
  \city{Changsha}
  \country{China}
}

\author{Weihao Gan}
\authornotemark[2]
\email{ganweihao@mlslabs.com.cn}
\affiliation{%
  \institution{Malanshan Audio\&Video Laboratory}
  \city{Changsha}
  \country{China}
}

\author{Zixing Zhang}
\authornotemark[2]
\email{zixingzhang@hnu.edu.cn}
\affiliation{%
  \institution{Hunan University}
  \city{Changsha}
  \country{China}
}
\renewcommand{\shortauthors}{Guo et al.}

\begin{abstract}
Despite extensive research on textual and visual disambiguation, disambiguation through speech (DTS) remains underexplored. This is largely due to the lack of high-quality datasets that pair spoken sentences with richly ambiguous text. To address this gap, we present DEBATE, a unique public Chinese speech-text dataset designed to study how speech cues and patterns—pronunciation, pause, stress and intonation—can help resolve textual ambiguity and reveal a speaker’s true intent. DEBATE contains 1,001 carefully selected ambiguous utterances, each recorded by 10 native speakers, capturing diverse linguistic ambiguities and their disambiguation through speech. We detail the data collection pipeline and provide rigorous quality analysis. Additionally, we benchmark three state-of-the-art large speech and audio-language models, illustrating clear and huge performance gaps between machine and human understanding of spoken intent. DEBATE represents the first effort of its kind and offers a foundation for building similar DTS datasets across languages and cultures. The dataset and associated code are available at: https://github.com/SmileHnu/DEBATE.
\end{abstract}


\begin{teaserfigure}
  \includegraphics[width=\linewidth, clip, trim=.9cm .6cm .3cm .1cm]{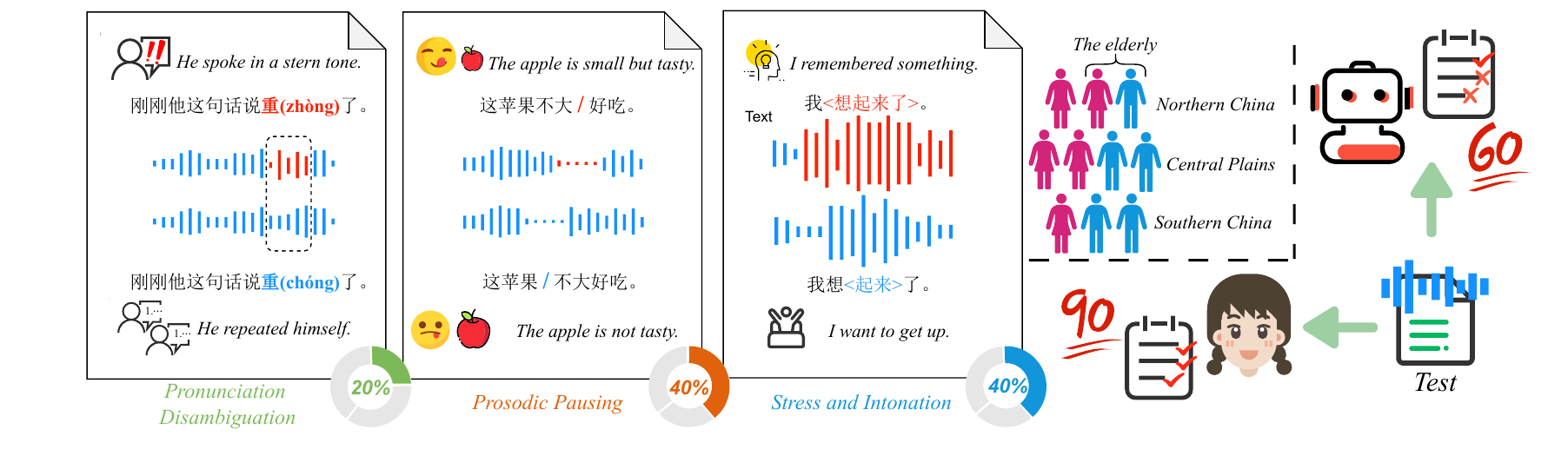}
  \caption{\textbf{DEBATE is the first multimodal dataset designed for speech-based disambiguation in Mandarin Chinese, focusing on how acoustic cues help resolve textual ambiguity.} The dataset comprises over 10K audio recordings of 1,001 ambiguous text prompts, each spoken by 10 native speakers and annotated with disambiguated meanings. The left figure illustrates three disambiguation scenarios.
  The right figure presents key dataset statistics, including speaker count, gender ratio, and scenario distribution. 
  By evaluating on DEBATE, significant performance gaps between large speech-language models and humans are noticed.
  }
  \Description{}
  \label{fig:example}
\end{teaserfigure}


\maketitle
\pagestyle{empty}

\input{sections/intro}
\input{sections/related}
\input{sections/dataset}
\input{sections/benchmarking}
\input{sections/conclusion}

\begin{acks}

This work was mainly supported by the Malanshan Audio\&Video Laboratory, China, and was partially supported by the Guangdong Basic and Applied Basic Research Foundation under Grant Nr.~2024A1515010112 and the Changsha Science and Technology Bureau Foundation under Grant Nr.~kq2402082.
We would like to thank all the native speakers who participated in the data collection and human inference process for their time and effort. Their support was essential in creating the DEBATE dataset.
\end{acks}





\balance
\bibliographystyle{ACM-Reference-Format}
\bibliography{ref}


\end{CJK}
\end{document}

%% file: sections/intro.tex
\section{Introduction}
\label{sec:intro}

\begin{center}
\textit{``Words mean more than what is set down on paper. 
\\It takes the human voice to infuse them with deeper meaning.''\\\hfill — Maya Angelou }

\noindent\textit{``听其言，观其音，察其意。''  — 汉$\cdot$班固} 
\end{center}
Despite advances in Large Language Models (LLMs), certain forms of ambiguity remain challenging to resolve~\cite{liu2023we, kamath2024scope}.
\textbf{Ambiguity arises naturally in human language} when words or phrases carry multiple possible interpretations.
For example, Mandarin Chinese, an isolating (analytic) language, relies primarily on semantic context to convey grammatical relationships~\cite{li1989mandarin, ross2024modern}.
This is different from synthetic languages, which leverage morphological inflections (e.\,g., changes in word form to indicate tense, number, or case) and use more complex grammatical conjugation~\cite{bickel2007inflectional, szemerenyi1999introduction}.
Therefore, while the structural flexibility of Mandarin enhances expressive richness, it also makes the language more prone to syntactic and semantic ambiguities, especially in written, text-only form~\cite{wang2024morpheme, zhang2024chambi}. Also, in Mandarin, words can be formed from either single or multiple characters. Unlike English, Chinese sentences lack explicit word boundaries, such as spaces, and rely primarily on limited punctuation marks to complete full sentences. For instance, a string like ``ABC'' (e.\,g.,``地面积'' ) can be interpreted as either ``AB/C'' (e.\,g.,``地面/积'' ) or ``A/BC''(e.\,g.,``地/面积'' ), both yielding valid but distinct meanings. This phenomenon is comparable to the classic English example ``Godisnowhere'', which can be understood as either ``God is now here'' or ``God is nowhere''. 

While such ambiguity can sometimes be a source of humour or wordplay, most real-world communication demands clarity and precision to avoid misunderstanding~\cite{bucaria2004lexical, majambere2011clarity}. Disambiguation has long been a central research topic in Chinese NLP~\cite{bevilacqua2021recent, wang-etal-2024-disambiguate, zheng2021leveraging}. A variety of text-based corpora have been developed,  and computational models have been proposed to resolve these ambiguities through techniques such as sentence rewriting or contextual reasoning~\cite{zheng2021leveraging, zhang2024chambi, wang2024morpheme, chen2014unified, hou2020try}. However, most existing efforts focus solely on textual input, ignoring other communicative modalities that humans naturally rely on for clarity.

Meanwhile, recent advancements in speech and audio understanding and their integration with LLMs have enabled audio-based understanding to play a more significant role in multimodal AI systems~\cite{deshmukh2023pengi,chu2023qwen, gonglisten, ghosh-etal-2024-gama,zhang2024refashioning}. In natural spoken communication, some of the ambiguities present in written Chinese can be effortlessly resolved. We categorise such speech-based disambiguation cues into three main types: (1) \textit{Pronunciation Disambiguation}: Certain characters in Chinese have multiple pronunciations, which can convey different meanings. When spoken aloud, the sound naturally disambiguates the intended word.
(2) \textit{Prosodic Pausing}: Although written Chinese does not use spaces to separate words, native speakers tend to pause between semantically distinct word groups during speech, providing cues for proper segmentation and interpretation. (3) \textit{Stress and Intonation}: Stress patterns are not represented in written text, but speakers may use stress and intonation to highlight intent and clarify meaning during oral communication.

Despite the growing availability of text-only ambiguity corpora and various speech-text paired datasets for ASR or question answering tasks, to the best of our knowledge, there is no existing dataset specifically designed to study how speech cues naturally disambiguate textual ambiguity in Mandarin. To address this gap, we introduce \textbf{DEBATE}: a novel speech-text corpus aimed at \textbf{D}isentangling T\textbf{e}xtual Am\textbf{b}iguity in M\textbf{a}ndarin \textbf{t}hrough Sp\textbf{e}ech. The DEBATE dataset is designed to evaluate and train models on their ability of disambiguation through speech (DTS). Our main contributions are as follows:\\

\begin{itemize}[noitemsep, topsep=0pt]
    \item We present DEBATE, a publicly available Mandarin speech-text corpus specifically designed to facilitate research on disambiguation through speech. The corpus includes \textit{1\,k+} text entries paired with \textit{10\,k+} audio samples (\textit{9.66} hours). 
    \item We established a structured data collection framework, outlining each step involved in constructing the database, including the development of the textual corpus and audio recording setups, to ensure both diversity and quality.
    \item We benchmark three large speech-language models. Our results show that these models still struggle to effectively leverage acoustic information for understanding true intent, performing much worse than human participants on DTS.
\end{itemize}

%% file: sections/related.tex
\section{Related Work}
\label{sec:related}

In recent years, several corpora have been developed for solving disambiguation problems in Chinese text. Some focus on the character level, identifying commonly used morphemes with multiple interpretations and building corresponding vocabularies~\cite{wang2024morpheme}. Others operate at the sentence level, crawling from the web data or integrating specialised corpora to create Word Sense Disambiguation~(WSD) datasets~\cite{zhang2024chambi, yan-etal-2023-ji}. Similarly, many efforts have been made in other languages and cultures. Well-known examples of such efforts can be the multilingual WSD tasks from the SemEval evaluation series~\cite{navigli-etal-2013-semeval, moro-navigli-2015-semeval}. These works provide valuable foundations for understanding the disambiguation issue and inspire the development of our own text corpus.

Moreover, various speech and multimodal datasets have been proposed to advance spoken language understanding, particularly for uncovering nuanced speaker intent beyond the literal meaning of words~\cite{wu2021mumor, bedi2021multi, park2016multimodal, zhang2024paralbench}. For example, some datasets focus on identifying the intended recipient of an utterance in multi-party conversations~\cite{schuller2017interspeech, tesema2023addressee, akhtiamov2017you}, while others focus on detecting humour and sarcasm~\cite{bedi2021multi, yue-etal-2024-sarcnet, christ2024towards}. These studies highlighted that prosodic features such as stress and intonation play a vital role in disambiguating speech and inferring the true intent in real-life communication~\cite{cheang2008sound}. Additionally, dialogue datasets have been constructed to build models in enabling more natural and effective communication~\cite{rastogi2020towards, li2023diversifying}. 

While these resources have largely advanced research in speech-based understanding, none have explicitly tackled the challenge of DTS, especially for Mandarin. The absence of a high-quality dataset has thus hindered progress in this area. To bridge this gap, our work introduces a novel dataset specifically designed to investigate how spoken cues, such as pronunciation, pauses, and stress, can aid in disambiguating semantically ambiguous text in Mandarin Chinese.

%% file: sections/dataset.tex
\section{DEBATE Dataset}
\label{sec:dataset}
\subsection{Dataset Overview}


The issue of ambiguity in Chinese has long existed, drawing widespread attention in linguistic research and posing persistent challenges for natural language processing systems. Depending on the modality through which information is conveyed, such ambiguities can be broadly divided into two categories: those that arise in the spoken modality but can be resolved using textual cues, and those that exist in text but require speech-based signals—such as pronunciation, intonation, and pauses—for disambiguation. The DEBATE dataset we constructed focuses primarily on the latter, where surface-level textual content is ambiguous, yet the intended meaning can be clarified through prosodic and phonetic information through speech.

Specifically, DEBATE covers three representative types of speech-assisted disambiguation scenarios. The three scenarios are categorised based on the primary speech cues required to resolve ambiguity, as outlined below.

First, \textbf{polyphonic character ambiguity} is a common phenomenon in Chinese, where a single character may have multiple pronunciations and corresponding meanings. This often results in textual expressions that are inherently ambiguous or even appear as puns. For instance, in the left example of Figure~\ref{fig:example}, the character ``重'' could be interpreted as indicating either a serious tone or repeated speech, depending on its pronunciation. In written form without sufficient context, determining the intended meaning is difficult. However, the speech signal clearly reveals the correct pronunciation, allowing the listener (and ideally a multimodal model) to resolve the ambiguity accurately.

Second, \textbf{structural ambiguity} arises due to the absence of explicit word boundaries and punctuation in Chinese text. Different syntactic segmentations may lead to entirely different meanings. As shown in the middle example of Figure ~\ref{fig:example}, the same sentence can be parsed to suggest that the apple is tasty or not, depending on how it is segmented. In spoken language, the speaker’s natural prosodic pauses provide crucial structural cues that indicate sentence boundaries. By leveraging features such as the position and duration of pauses, one can infer sentence structure and reduce ambiguity at the syntactic level.

Third, ambiguity can stem from differences in semantic focus, which are often marked by \textbf{stress or intonation ambiguity}. In Chinese, shifting the position of stress within a sentence can significantly alter its meaning. For example, in the third pair of sentences shown in Figure ~\ref{fig:example}, when the stress falls on ``想起来了'', it indicates that the speaker has recalled something. In contrast, when the stress is placed on ``起来'' , it suggests an intention to physically get up. Such semantic ambiguities caused by stress placement are often difficult to detect in text, but in spoken communication, speakers typically clarify the intended meaning through variations in pitch or intensity. 

In summary, DEBATE targets ambiguity types that are textually obscure yet resolvable through speech.

\begin{figure*}[t]  
    \centering
    \includegraphics[width=.9\textwidth,clip, trim=.9cm .4cm .7cm .1cm]{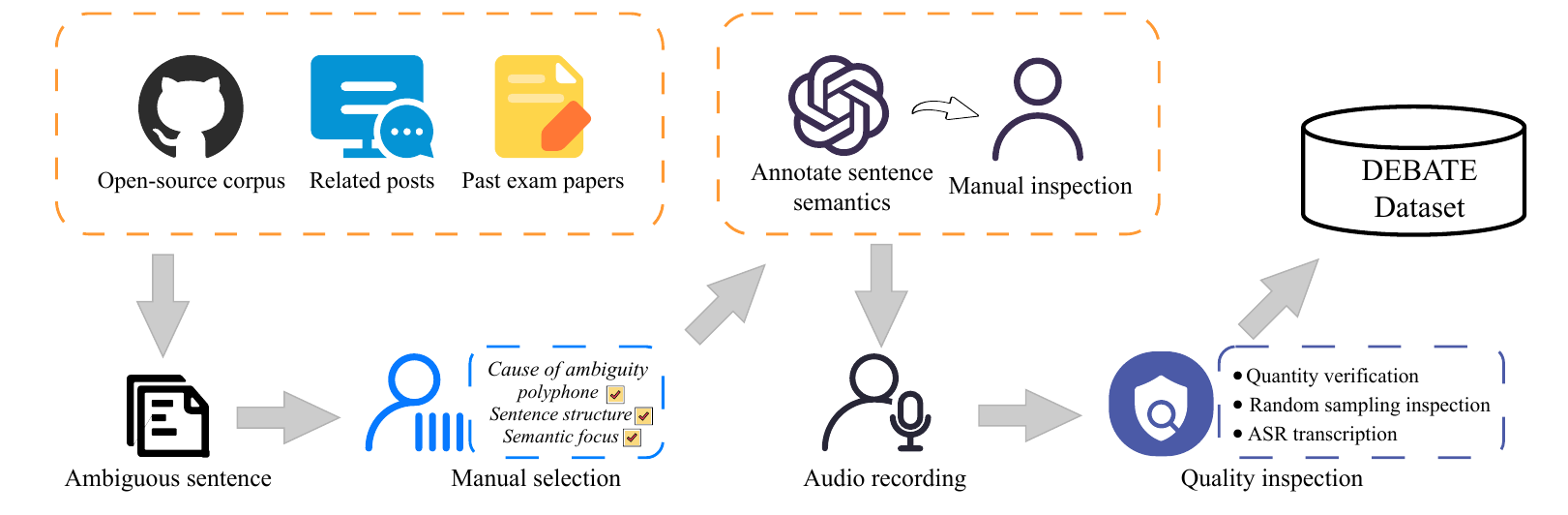}
    \caption{\textbf{Pipeline for creating DEBATE—the first speech-text dataset designed for Chinese disambiguation through speech.} Ambiguous text was initially collected from diverse sources and manually inspected to identify suitable candidate sentences. These sentences were then augmented via LLMs and underwent 2nd round of manual validation. Then, selected sentences were read by 10 native speakers, and went through rigorous quality control. The resulting DEBATE dataset includes over 10k curated speech recordings with rich annotations.
}
    \label{fig:data_collect}
\end{figure*}

\begin{figure}[t]
    \centering
    \includegraphics[width=\columnwidth,clip, trim=.3cm .3cm .3cm .1cm]{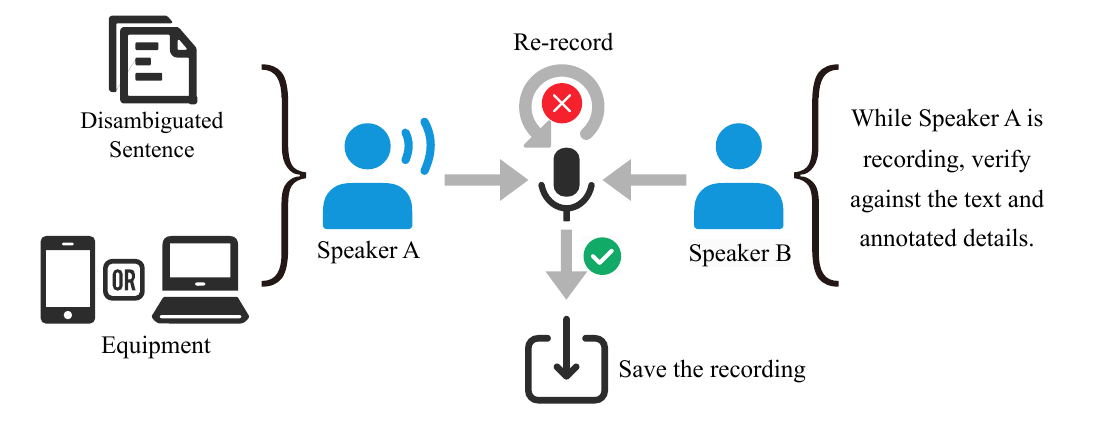}
    \caption{\textbf{Illustration of the audio recording process.} Each session involves two participants: one dedicated speaker and one passive listener, recorded in parallel to maximise recording quality.}
    \label{fig:record}
\end{figure}


\subsection{Data Generation Pipeline}
The overall pipeline is illustrated in Figure~\ref{fig:data_collect}, comprising three main stages. We depict each stage in the following.

\subsubsection{Raw Text Gathering and Annotation}\hfill

We first constructed a Chinese text dataset tailored for research on speech-based disambiguation.  The data was collected from three primary sources: (1) open-source corpora, (2) social media platforms, and (3) standardised examination question banks.

Specifically, we extracted representative ambiguous samples directly from publicly available Chinese ambiguity corpora hosted on GitHub. For social media data, we retrieved user-generated posts exhibiting semantic ambiguity by querying with keywords such as ``ambiguity'', ``phrase segmentation'', and ``stress''. Additionally, we selected structurally complex and ambiguous examples from sentence analysis tasks in the verbal reasoning sections of Chinese civil service examinations.

After collecting the text, we performed systematic manual annotation to classify the data into three distinct types of ambiguity: \textit{polyphonic character ambiguity}, \textit{structural ambiguity}, and \textit{focus ambiguity}. Samples exhibiting other types of semantic ambiguity outside these categories were excluded. To further enrich the dataset in terms of coverage and syntactic diversity, we employed large language models (LLMs)  such as GPT and DeepSeek to generate additional sentences following the structure and style of the manually collected examples. All machine-generated samples underwent human verification to ensure grammaticality, logical consistency, and alignment with the target ambiguity type.

To maintain ethical standards, all sentences containing inappropriate or sensitive content were removed to ensure the dataset's compliance and public usability. As a result of this process, we curated a total of \textit{200} samples of polyphonic character ambiguity, \textit{401} ones with structural ambiguity, and \textit{400} ones of focus ambiguity. 

After collecting and preliminarily filtering the ambiguous text, we further introduced three prosodic-level strategies to systematically disambiguate the selected sentences. These strategies specifically target three common types of ambiguity: polyphonic character ambiguity, structural ambiguity, and focus-related ambiguity.

\textit{For polyphonic character ambiguity}, we explicitly annotated the pronunciation of each character to eliminate semantic confusion caused by identical written forms with different pronunciations.

\textit{For structural ambiguity}, we incorporated prosodic annotation by using the ``/'' symbol to mark prosodic boundaries within sentences. These boundaries indicate appropriate short pauses in speech, which help clarify syntactic structures and enhance sentence intelligibility.

\textit{For focus ambiguity}, we marked key words requiring emphasis with stress symbols or pitch-rise indicators. This approach reinforces the pragmatic focus of the sentence and assists listeners in accurately identifying the speaker’s intended meaning and communicative emphasis.

To further enhance the usability of this dataset in semantic understanding and downstream evaluation tasks, we generated semantic annotations for each sentence. These annotations were produced using LLMs,
generating multiple candidate explanations per sentence. We then conducted human review to evaluate the fluency, semantic accuracy, and logical consistency of each candidate. Only the most optimal explanation per disambiguated sentence was retained. 


\subsubsection{Speech Data Recording}\hfill

To ensure both diversity and quality control of the speech data, we recruited ten volunteers to participate in the audio recording task, including eight young adults and two elderly individuals, with a balanced gender ratio (5:5) to ensure demographic representativeness in terms of both age and gender. All participants received detailed instructions prior to recording, including specific guidance on prosodic features such as stress placement, intonation patterns, and appropriate pause locations based on different task types.

The recordings were conducted using participants’ own devices, primarily consisting of smartphone microphones and built-in laptop microphones. This flexible device setup was intentionally adopted to closely mimic real-world deployment scenarios of speech models, thereby enhancing the ecological validity and generalisability of the dataset. To control recording quality, each participant was required to submit ten test recordings prior to the formal session. These samples were manually reviewed to evaluate recording equipment performance and environmental conditions, such as background noise and reverberation levels. Participants who met the quality standards were allowed to proceed to the full recording task.

During the formal recording sessions, we implemented a two-person collaborative mechanism: one individual performed the recording while the other monitored in real time, listening for issues such as mispronunciations, omissions, or semantic inconsistencies (see Fig.~\ref{fig:record}). If any issues were noticed, immediate feedback was provided and the utterance was re-recorded. This approach helped improve both the accuracy and consistency of the recordings, effectively preventing the accumulation of low-quality data.

As a result, we constructed a high-quality, multi-speaker speech dataset that maintains both naturalness and precision. 


\subsubsection{Quality Control}\hfill

Once the recording phase was completed, we conducted a systematic post-review of all recorded materials to ensure corpus completeness and audio quality, minimising issues such as missing files, duplicates, or speech defects. First, we verified the one-to-one correspondence between text entries and audio files, checking for potential omissions or redundant recordings to guarantee that each text sample was matched with a unique audio file per speaker.

Building on this, we performed a quality assessment of each speaker’s recordings across different tasks. Specifically, ten audio samples were randomly selected from each task and manually evaluated for pronunciation clarity, sentence completeness, naturalness of speech rate, and adherence to prosodic norms such as stress placement and pausing. This strategy enabled efficient identification of potential quality concerns and allowed for timely corrective actions. The sampling results indicated that all recordings met the expected standards, demonstrating overall high quality.

To further quantify audio quality, we exploited two advanced automatic speech recognition (ASR) systems, i.\,e., Whisper-large-v3-turbo~\cite{radford2023robust} and SenseVoice-small~\cite{an2024funaudiollm} to transcribe all recordings and compute Character Error Rate (CER) as a metric of alignment between audio and reference transcripts. The evaluation results are presented in Table~\ref{tab:cer_models}. Notably, sensevoice-small achieves CER of 2.96\% and 3.80\% on the two high-quality Mandarin speech recognition corpora AISHELL-1~\cite{bu2017aishell} test and AISHELL-2~\cite{du2018aishell} test\_ios, respectively, while attaining CERs of 2.82\% and 1.94\% on the $T_{Pause}$ and $T_{Stres}$ datasets. This performance demonstrates the high consistency between audio and text in our DEBATE dataset.

\begin{table}[t]
\centering
\caption{\textbf{Character error rate (\%) of ASR models on DEBATE.}}
\resizebox{\columnwidth}{!}{%
\begin{tabular}{lcccc}
\toprule
Models & $T_{Proun}$ & $T_{Pause}$ & $T_{Stres}$ & Average \\
\midrule
Whisper-large-v3 turbo & 9.66 & 7.14 & 5.28 & 7.37 \\
SenseVoice-Small & 4.75 & 2.82 & 1.94 & 3.48  \\
\bottomrule
\end{tabular}
}
\label{tab:cer_models}
\end{table}

\subsection{Dataset Statistics}
In total, the DEBATE dataset comprises approximately 9.66 hours of high-quality speech data spanning a variety of disambiguation task scenarios. Detailed descriptions of dataset structure, task distribution, and quality evaluation metrics can be found in Table~\ref{tab:audio_stats}. In particular, individual recordings range from 1.15 seconds to 11.80 seconds in duration, and the corpus exhibits strong diversity and representativeness in content. These characteristics make DEBATE a robust and valuable asset for future research and applications.


\begin{table}[t]
\centering
\caption{\textbf{Detailed statistics of the DEBATE dataset.}}
\begin{tabular}
{@{}p{1.1cm}p{1.3cm}p{1cm}p{1.5cm}p{1.8cm}@{}}
\toprule
\textbf{Tasks} & \textbf{\#Samples} & \textbf{Hours} & \textbf{Mean Duration (s)} & \textbf{Duration Range (s)} \\
\midrule
$T_{Proun}$   & 2\,000  & 1.64  & 2.94   & 1.15-5.80  \\
$T_{Pause}$   & 4\,010  & 4.28  & 3.84  & 1.60-11.80  \\
$T_{Stres}$   & 4\,000  & 3.74  & 3.37  & 1.43-8.51   \\
\midrule
Total       & 10\,010  & 9.66 & 3.47 & 1.15-11.80  \\
\bottomrule
\end{tabular}
\label{tab:audio_stats}
\end{table}


%% file: sections/benchmarking.tex
\section{Benchmarks and Results}
\label{sec:exper}

As a showcase of speech-based disambiguation using the DEBATE dataset, we benchmarked several representative Large Speech Language Models (LSLMs). The primary objective is to assess how well these models can interpret semantically ambiguous sentences by leveraging speech cues. Specifically, we examine their ability to utilise salient acoustic features—such as pronunciation differences, prosodic boundaries, speech rate and rhythm, and stress patterns—to resolve ambiguities stemming from polyphonic characters, syntactic structures, and unclear semantic focus within text.

\subsection{Experimental Setup}
For this aim, we selected three widely-used LSLMs for evaluation, i.\,e., Qwen2-audio~\cite{chu2023qwen}, Qwen2.5-omni~\cite{xu2025qwen2}, and Gemini 2.0 Flash~\cite{gemini24}. These models have demonstrated excellent performance in speech understanding tasks. All evaluations were conducted under a zero-shot inference setting to examine whether the models can perform semantic disambiguation.

For task construction, 
speech samples were used as input queries, paired with carefully designed text prompts to specify the semantic discrimination objective. Each test item was formatted as a single-choice question: among multiple plausible interpretations of an ambiguous sentence, two were manually selected as answer options, only one of which matched the intended meaning conveyed in the audio. The model was required to select the correct interpretation based on the provided audio. The detailed prompt template used for constructing these tasks is provided in Table~\ref {tab:experiment_prompt}.

\begin{table}[t]
\centering
\caption{\textbf{Prompt construction templates used for model inference tasks.} Text in parentheses is for illustrative purposes only and not included in the actual prompts. During each inference run, only one of three alternative text blocks (highlighted with background colours) is selected and used. Note that, original prompts were written in Chinese and are translated into English here for clarity and presentation.}
\begin{tabular}{@{}p{0.95\linewidth}@{}}
\toprule
\textit{\textbf{Prompt Template}}\\
\midrule
As a model equipped with professional-level audio semantic understanding capabilities, you are expected to accurately identify the precise meaning conveyed in the given audio segment.\\

\addlinespace[0.25em]
{\cellcolor{green!20}(\textit{$T_{Proun}$ only}) A core challenge lies in handling polyphonic characters. You must correctly identify the actual pronunciation of each polyphonic word within the audio and, by integrating contextual information, accurately infer its intended meaning—an essential step for overall semantic interpretation. }\\

\addlinespace[0.25em]
\cellcolor{orange!20}(\textit{$T_{Pause}$ only}) A core aspect of this task lies in analysing pause-related features—such as the duration of silence and prosodic interruptions—which serve as primary cues for determining the hierarchical structure of the sentence. You must incorporate this structural information to accurately interpret the overall meaning and logical relationships within the sentence. \\

\addlinespace[0.25em]
\cellcolor{blue!20}(\textit{$T_{Stress}$ only}) A key challenge lies in capturing prosodic variations in the audio—such as pitch rises or falls, changes in speech rate, and stress patterns—which are essential cues for inferring the speaker's intended emphasis. \\

\addlinespace[0.25em]
I will provide two possible meanings for you to choose from:\\
A. [Meaning 1] \hspace{1cm} B. [Meaning 2] \\
Please choose the option you consider most appropriate. No other output text/explanation. Only provide the option.\\

\bottomrule
\end{tabular}
\label{tab:experiment_prompt}
\end{table}

\subsection{Results and Analysis}

\begin{table*}[t]
\centering
\caption{\textbf{Evaluation of large-scale language models (LSLMs) on the DEBATE dataset in terms of accuracy~(\%) and macro-F1 score~(\%).} Results are reported for three models: Qwen2-Audio, Qwen2.5-Omni, and Gemini 2.0 Flash. For each task, we present the mean and variance across 10 speakers. The best performance in each column is highlighted.}
\begin{tabular}{lcccccc}
\toprule
Models & \multicolumn{2}{c}{$\mathit{T_{Proun}}$} & \multicolumn{2}{c}{$\mathit{T_{Pause}}$} & \multicolumn{2}{c}{$\mathit{T_{Stres}}$} \\
\cmidrule(lr){2-3} \cmidrule(lr){4-5} \cmidrule(lr){6-7}
& Accuracy & F1 & Accuracy & F1 & Accuracy & F1 \\



\textit{Qwen2-Audio}       & 58.60 $\pm$3.43 & 55.40 $\pm$4.03 & 55.64 $\pm$1.85   & 51.10 $\pm$ 1.66 & 51.57 $\pm$1.01 & 47.60 $\pm$1.43 \\

\textit{Qwen2.5-Omni}       & \textbf{65.65 $\pm$2.30} & \textbf{65.20 $\pm$2.62} & \textbf{68.08 $\pm$1.47} & \textbf{67.90 $\pm$1.60} & 55.02 $\pm$2.16 & 55.20 $\pm$2.15 \\

\textit{Gemini 2.0 Flash}   & 61.15 $\pm$3.95 & 60.80 $\pm$4.16 & 68.00 $\pm$1.69 & 67.90$\pm$1.79 & \textbf{58.83 $\pm$4.06} & \textbf{58.00 $\pm$4.47} \\

\bottomrule
\end{tabular}
\label{tab:results}
\end{table*}


We report performance in terms of accuracy and macro-F1.
 Table~\ref{tab:results} summarises the performance of three selected LSLMs across different types of semantic disambiguation tasks. Overall, Gemini-2.0-Flash and Qwen2.5-Omni demonstrated comparatively strong performance, exhibiting competitive capabilities across tasks, whereas Qwen2-Audio showed relatively weaker results. This performance gap can be partially attributed to the more powerful underlying LLM architectures and the larger-scale training corpora used by the former two models.

In terms of specific task performance, all three models demonstrated generally decent results in polyphone disambiguation ($T_{Proun}$) and sentence structure understanding tasks ($T_{Pause}$). Notably, in the sentence structure task, the model outputs were more stable and consistent, indicating that current LSLMs possess the foundational capabilities to detect intra-sentence pauses. This suggests that these models can, to a certain extent, perceive hierarchical sentence structures and infer content that aligns more closely with the intended semantic meaning.

However, in the stress understanding task, all models performed poorly, with Qwen2-Audio in particular producing results that were nearly random. This highlights a clear limitation in the models’ ability to capture finer prosodic features in speech, such as stress patterns and pitch variations. Compared to pauses, which are relatively explicit and have clear temporal boundaries, stress involves more complex acoustic variations within the speech signal. Accurate recognition of speaker emphasis at the semantic level thus requires a higher degree of prosodic modelling capability. Moreover, current pre-trained LSLMs are predominantly trained on large-scale, automatically transcribed corpora, in which prosodic features like stress are rarely annotated. This further constrains the models’ ability to learn and generalise such nuanced characteristics.

\begin{figure}[t]
    \centering
    \includegraphics[width=\columnwidth]{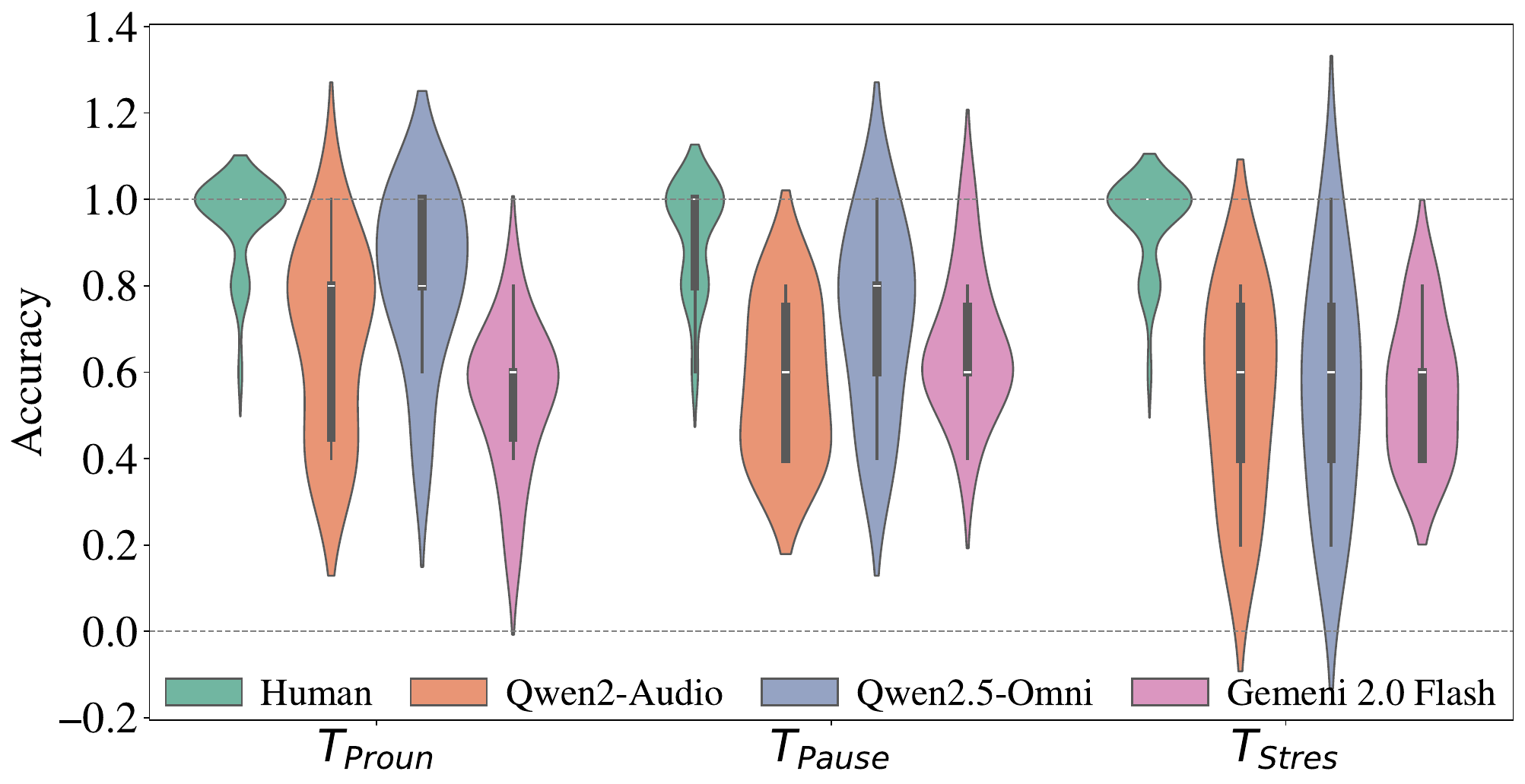}
    
    \caption{\textbf{Comparison of human and model performance on the small-scale test set with violin plot.} `Human' scores are the combined results from assessments made by three volunteers. }
    \vspace{-1cm}
    \label{fig:human_model}
\end{figure}

\textbf{{Comparing Human Performance and LSLMs.}} We randomly selected 50 samples per subtask of the DEBATE dataset to construct a small-scale evaluation set, ensuring that each speaker contributed five unique samples. This set was used for manual evaluation, and we also assessed the performance of LSLMs on the same set to highlight the gap between model and human capabilities in speech understanding. The evaluation procedure was consistent with the model inference process and was independently conducted by three volunteers who were not involved in this study. 
The violin plots in Fig.~\ref{fig:human_model} illustrate that LSLMs not only underperform compared to human listeners across all three tasks, but also exhibit greater performance dispersion. This suggests that all tested LSLMs struggle both in accuracy and in consistency across speakers and sentences.

In summary, while current LSLMs show potentials in perceiving certain speech cues, there remains substantial room for improvement. 
This highlights the critical need for future research to enhance models’ ability to interpret the rich prosodic and acoustic cues in speech—an essential step toward achieving deeper semantic understanding and more accurate user intent interpretation.

%% file: sections/conclusion.tex
\section{Conclusion}
\label{sec:conclusion}

We introduce DEBATE, a publicly available Mandarin speech-text corpus specifically designed for semantic disambiguation tasks. 
Using DEBATE, we conducted a systematic assessment of current large-scale speech-language models (LSLMs), including representative models from the Qwen and Gemini families. 
Our results suggest that current LSLMs exhibit a modest ability to perform speech-based disambiguation, with best accuracy about 58-68\% across all tasks and models—still well below human performance.
This limitation hinders their ability to accurately resolve complex semantic ambiguities, highlighting the need for further improvements in acoustic modelling within LSLMs. In addition to serving as a benchmark for semantic understanding, DEBATE also holds promise as a fine-tuning resource for text-to-speech (TTS) systems—particularly for improving pronunciation disambiguation of homographs and the natural rendering of prosodic stress. These remain key challenges in most current TTS systems, and the DEBATE corpus offers targeted data to address them. Moreover, the dataset contains implicit regional phonetic variation, making it a valuable resource for modelling linguistic variants. This opens up possibilities for research in sociolinguistics and language adaptation, especially in studying the regional adaptability and standardisation of Mandarin pronunciation across different dialectal backgrounds.


\textbf{Limitations} The current version of DEBATE covers only Mandarin Chinese. Besides, we evaluated only three existing models, which, while representative, may not reflect the full spectrum of approaches in audio-language understanding. Future work should explore broader model families and training paradigms.


\textbf{Ethics Statement} All data collection was conducted with informed consent from native speakers. Individuals cannot be identified or re-identified from the data to protect user privacy. The dataset is intended strictly for academic and research purposes. Care was taken to ensure diversity in audio/text samples and to avoid any content that could be considered harmful or offensive.

%% file: main.bbl

\begin{thebibliography}{41}


\ifx \showCODEN    \undefined \def \showCODEN     #1{\unskip}     \fi
\ifx \showISBNx    \undefined \def \showISBNx     #1{\unskip}     \fi
\ifx \showISBNxiii \undefined \def \showISBNxiii  #1{\unskip}     \fi
\ifx \showISSN     \undefined \def \showISSN      #1{\unskip}     \fi
\ifx \showLCCN     \undefined \def \showLCCN      #1{\unskip}     \fi
\ifx \shownote     \undefined \def \shownote      #1{#1}          \fi
\ifx \showarticletitle \undefined \def \showarticletitle #1{#1}   \fi
\ifx \showURL      \undefined \def \showURL       {\relax}        \fi
\providecommand\bibfield[2]{#2}
\providecommand\bibinfo[2]{#2}
\providecommand\natexlab[1]{#1}
\providecommand\showeprint[2][]{arXiv:#2}

\bibitem[Akhtiamov et~al\mbox{.}(2017)]%
        {akhtiamov2017you}
\bibfield{author}{\bibinfo{person}{Oleg Akhtiamov}, \bibinfo{person}{Dmitrii Ubskii}, \bibinfo{person}{Evgeniia Feldina}, \bibinfo{person}{Aleksei Pugachev}, \bibinfo{person}{Alexey Karpov}, {and} \bibinfo{person}{Wolfgang Minker}.} \bibinfo{year}{2017}\natexlab{}.
\newblock \showarticletitle{Are you addressing me? Multimodal addressee detection in human-human-computer conversations}. In \bibinfo{booktitle}{\emph{Proc.\ 19th International Conference on Speech and Computer (SPECOM)}}. \bibinfo{address}{Hatfield, UK}, \bibinfo{pages}{152--161}.
\newblock


\bibitem[An et~al\mbox{.}(2024)]%
        {an2024funaudiollm}
\bibfield{author}{\bibinfo{person}{Keyu An}, \bibinfo{person}{Qian Chen}, \bibinfo{person}{Chong Deng}, \bibinfo{person}{Zhihao Du}, \bibinfo{person}{Changfeng Gao}, \bibinfo{person}{Zhifu Gao}, {et~al\mbox{.}}} \bibinfo{year}{2024}\natexlab{}.
\newblock \showarticletitle{Fun{A}udio{LLM}: Voice understanding and generation foundation models for natural interaction between humans and {LLMs}}.
\newblock \bibinfo{journal}{\emph{arXiv preprint arXiv:2407.04051}} (\bibinfo{year}{2024}).
\newblock


\bibitem[Bedi et~al\mbox{.}(2021)]%
        {bedi2021multi}
\bibfield{author}{\bibinfo{person}{Manjot Bedi}, \bibinfo{person}{Shivani Kumar}, \bibinfo{person}{Md~Shad Akhtar}, {and} \bibinfo{person}{Tanmoy Chakraborty}.} \bibinfo{year}{2021}\natexlab{}.
\newblock \showarticletitle{Multi-modal sarcasm detection and humor classification in code-mixed conversations}.
\newblock \bibinfo{journal}{\emph{IEEE Transactions on Affective Computing}} \bibinfo{volume}{14}, \bibinfo{number}{2} (\bibinfo{year}{2021}), \bibinfo{pages}{1363--1375}.
\newblock


\bibitem[Bevilacqua et~al\mbox{.}(2021)]%
        {bevilacqua2021recent}
\bibfield{author}{\bibinfo{person}{Michele Bevilacqua}, \bibinfo{person}{Tommaso Pasini}, \bibinfo{person}{Alessandro Raganato}, {and} \bibinfo{person}{Roberto Navigli}.} \bibinfo{year}{2021}\natexlab{}.
\newblock \showarticletitle{Recent trends in word sense disambiguation: A survey}. In \bibinfo{booktitle}{\emph{Proc.\ International Joint Conference on Artificial Intelligence (IJCAI)}}. \bibinfo{address}{Montreal, Canada}, \bibinfo{pages}{4330--4338}.
\newblock


\bibitem[Bickel and Nichols(2007)]%
        {bickel2007inflectional}
\bibfield{author}{\bibinfo{person}{Balthasar Bickel} {and} \bibinfo{person}{Johanna Nichols}.} \bibinfo{year}{2007}\natexlab{}.
\newblock \showarticletitle{Inflectional morphology}.
\newblock \bibinfo{journal}{\emph{Language Typology and Syntactic Description}} \bibinfo{volume}{3}, \bibinfo{number}{2} (\bibinfo{year}{2007}), \bibinfo{pages}{169--240}.
\newblock


\bibitem[Bu et~al\mbox{.}(2017)]%
        {bu2017aishell}
\bibfield{author}{\bibinfo{person}{Hui Bu}, \bibinfo{person}{Jiayu Du}, \bibinfo{person}{Xingyu Na}, \bibinfo{person}{Bengu Wu}, {and} \bibinfo{person}{Hao Zheng}.} \bibinfo{year}{2017}\natexlab{}.
\newblock \showarticletitle{AISHELL-1: An open-source mandarin speech corpus and a speech recognition baseline}. In \bibinfo{booktitle}{\emph{Proc.\ 20th Conference of the Oriental Chapter of the International Coordinating Committee on Speech Databases and Speech I/O Systems and Assessment (O-COCOSDA)}}. \bibinfo{address}{Seoul, Korea}, \bibinfo{pages}{1--5}.
\newblock


\bibitem[Bucaria(2004)]%
        {bucaria2004lexical}
\bibfield{author}{\bibinfo{person}{Chiara Bucaria}.} \bibinfo{year}{2004}\natexlab{}.
\newblock \showarticletitle{Lexical and syntactic ambiguity as a source of humor: The case of newspaper headlines}.
\newblock \bibinfo{journal}{\emph{Humor}} \bibinfo{volume}{17}, \bibinfo{number}{3} (\bibinfo{year}{2004}), \bibinfo{pages}{279--309}.
\newblock


\bibitem[Cheang and Pell(2008)]%
        {cheang2008sound}
\bibfield{author}{\bibinfo{person}{Henry~S Cheang} {and} \bibinfo{person}{Marc~D Pell}.} \bibinfo{year}{2008}\natexlab{}.
\newblock \showarticletitle{The sound of sarcasm}.
\newblock \bibinfo{journal}{\emph{Speech Communication}} \bibinfo{volume}{50}, \bibinfo{number}{5} (\bibinfo{year}{2008}), \bibinfo{pages}{366--381}.
\newblock


\bibitem[Chen et~al\mbox{.}(2014)]%
        {chen2014unified}
\bibfield{author}{\bibinfo{person}{Xinxiong Chen}, \bibinfo{person}{Zhiyuan Liu}, {and} \bibinfo{person}{Maosong Sun}.} \bibinfo{year}{2014}\natexlab{}.
\newblock \showarticletitle{A unified model for word sense representation and disambiguation}. In \bibinfo{booktitle}{\emph{Proc.\ Conference on Empirical Methods in Natural Language Processing (EMNLP)}}. \bibinfo{address}{Doha, Qatar}, \bibinfo{pages}{1025--1035}.
\newblock


\bibitem[Christ et~al\mbox{.}(2025)]%
        {christ2024towards}
\bibfield{author}{\bibinfo{person}{Lukas Christ}, \bibinfo{person}{Shahin Amiriparian}, \bibinfo{person}{Alexander Kathan}, \bibinfo{person}{Niklas M{\"u}ller}, \bibinfo{person}{Andreas K{\"o}nig}, {and} \bibinfo{person}{Bj{\"o}rn~W Schuller}.} \bibinfo{year}{2025}\natexlab{}.
\newblock \showarticletitle{Towards multimodal prediction of spontaneous humor: A novel dataset and first results}.
\newblock \bibinfo{journal}{\emph{IEEE Transactions on Affective Computing}} \bibinfo{volume}{16}, \bibinfo{number}{2} (\bibinfo{year}{2025}), \bibinfo{pages}{844--860}.
\newblock


\bibitem[Chu et~al\mbox{.}(2023)]%
        {chu2023qwen}
\bibfield{author}{\bibinfo{person}{Yunfei Chu}, \bibinfo{person}{Jin Xu}, \bibinfo{person}{Xiaohuan Zhou}, \bibinfo{person}{Qian Yang}, \bibinfo{person}{Shiliang Zhang}, \bibinfo{person}{Zhijie Yan}, {et~al\mbox{.}}} \bibinfo{year}{2023}\natexlab{}.
\newblock \showarticletitle{Qwen-audio: Advancing universal audio understanding via unified large-scale audio-language models}.
\newblock \bibinfo{journal}{\emph{arXiv preprint arXiv:2311.07919}} (\bibinfo{year}{2023}).
\newblock


\bibitem[DeepMindg(2024)]%
        {gemini24}
\bibfield{author}{\bibinfo{person}{Google DeepMindg}.} \bibinfo{year}{2024}\natexlab{}.
\newblock \bibinfo{booktitle}{\emph{Introducing Gemini 2.0: our new AI model for the agentic era}}.
\newblock
\urldef\tempurl%
\url{https://blog.google/technology/google-deepmind/google-gemini-ai-update-december-2024/}
\showURL{%
Retrieved May 30, 2025 from \tempurl}


\bibitem[Deshmukh et~al\mbox{.}(2023)]%
        {deshmukh2023pengi}
\bibfield{author}{\bibinfo{person}{Soham Deshmukh}, \bibinfo{person}{Benjamin Elizalde}, \bibinfo{person}{Rita Singh}, {and} \bibinfo{person}{Huaming Wang}.} \bibinfo{year}{2023}\natexlab{}.
\newblock \showarticletitle{Pengi: An audio language model for audio tasks}. In \bibinfo{booktitle}{\emph{Proc.\ Advances in Neural Information Processing Systems (NeurIPS)}}, Vol.~\bibinfo{volume}{36}. \bibinfo{address}{New Orleans, LA}, \bibinfo{pages}{18090--18108}.
\newblock


\bibitem[Du et~al\mbox{.}(2018)]%
        {du2018aishell}
\bibfield{author}{\bibinfo{person}{Jiayu Du}, \bibinfo{person}{Xingyu Na}, \bibinfo{person}{Xuechen Liu}, {and} \bibinfo{person}{Hui Bu}.} \bibinfo{year}{2018}\natexlab{}.
\newblock \showarticletitle{AISHELL-2: Transforming mandarin {ASR} research into industrial scale}.
\newblock \bibinfo{journal}{\emph{arXiv preprint arXiv:1808.10583}} (\bibinfo{year}{2018}).
\newblock


\bibitem[Ghosh et~al\mbox{.}(2024)]%
        {ghosh-etal-2024-gama}
\bibfield{author}{\bibinfo{person}{Sreyan Ghosh}, \bibinfo{person}{Sonal Kumar}, \bibinfo{person}{Ashish Seth}, \bibinfo{person}{Chandra Kiran~Reddy Evuru}, \bibinfo{person}{Utkarsh Tyagi}, \bibinfo{person}{S Sakshi}, {et~al\mbox{.}}} \bibinfo{year}{2024}\natexlab{}.
\newblock \showarticletitle{{GAMA}: A large audio-language model with advanced audio understanding and complex reasoning abilities}. In \bibinfo{booktitle}{\emph{Proc.\ Conference on Empirical Methods in Natural Language Processing (EMNLP)}}. \bibinfo{address}{Miami, FL}, \bibinfo{pages}{6288--6313}.
\newblock


\bibitem[Gong et~al\mbox{.}(2024)]%
        {gonglisten}
\bibfield{author}{\bibinfo{person}{Yuan Gong}, \bibinfo{person}{Hongyin Luo}, \bibinfo{person}{Alexander~H Liu}, \bibinfo{person}{Leonid Karlinsky}, {and} \bibinfo{person}{James~R Glass}.} \bibinfo{year}{2024}\natexlab{}.
\newblock \showarticletitle{Listen, think, and understand}. In \bibinfo{booktitle}{\emph{Proc.\ 12th International Conference on Learning Representations (ICLR)}}. \bibinfo{address}{Vienna, Austria}, \bibinfo{pages}{30 pages}.
\newblock


\bibitem[Hou et~al\mbox{.}(2020)]%
        {hou2020try}
\bibfield{author}{\bibinfo{person}{Bairu Hou}, \bibinfo{person}{Fanchao Qi}, \bibinfo{person}{Yuan Zang}, \bibinfo{person}{Xurui Zhang}, \bibinfo{person}{Zhiyuan Liu}, {and} \bibinfo{person}{Maosong Sun}.} \bibinfo{year}{2020}\natexlab{}.
\newblock \showarticletitle{Try to substitute: An unsupervised {Chinese} word sense disambiguation method based on {Hownet}}. In \bibinfo{booktitle}{\emph{Proc.\ 28th International Conference on Computational Linguistics (COLING)}}. \bibinfo{address}{Barcelona, Spain}, \bibinfo{pages}{1752--1757}.
\newblock


\bibitem[Kamath et~al\mbox{.}(2024)]%
        {kamath2024scope}
\bibfield{author}{\bibinfo{person}{Gaurav Kamath}, \bibinfo{person}{Sebastian Schuster}, \bibinfo{person}{Sowmya Vajjala}, {and} \bibinfo{person}{Siva Reddy}.} \bibinfo{year}{2024}\natexlab{}.
\newblock \showarticletitle{Scope ambiguities in large language models}.
\newblock \bibinfo{journal}{\emph{Transactions of the Association for Computational Linguistics}}  \bibinfo{volume}{12} (\bibinfo{year}{2024}), \bibinfo{pages}{738--754}.
\newblock


\bibitem[Li et~al\mbox{.}(2023)]%
        {li2023diversifying}
\bibfield{author}{\bibinfo{person}{Bo Li}, \bibinfo{person}{Huan Zhao}, {and} \bibinfo{person}{Zixing Zhang}.} \bibinfo{year}{2023}\natexlab{}.
\newblock \showarticletitle{Diversifying emotional dialogue generation via selective adversarial training}.
\newblock \bibinfo{journal}{\emph{Sensors}} \bibinfo{volume}{23}, \bibinfo{number}{13}, Article \bibinfo{articleno}{5904} (\bibinfo{year}{2023}), \bibinfo{numpages}{16}~pages.
\newblock


\bibitem[Li and Thompson(1989)]%
        {li1989mandarin}
\bibfield{author}{\bibinfo{person}{Charles~N Li} {and} \bibinfo{person}{Sandra~A Thompson}.} \bibinfo{year}{1989}\natexlab{}.
\newblock \bibinfo{booktitle}{\emph{{Mandarin Chinese: A functional Reference Grammar}}}.
\newblock \bibinfo{publisher}{Univ of California Press}.
\newblock


\bibitem[Liu et~al\mbox{.}(2023)]%
        {liu2023we}
\bibfield{author}{\bibinfo{person}{Alisa Liu}, \bibinfo{person}{Zhaofeng Wu}, \bibinfo{person}{Julian Michael}, \bibinfo{person}{Alane Suhr}, \bibinfo{person}{Peter West}, \bibinfo{person}{Alexander Koller}, {et~al\mbox{.}}} \bibinfo{year}{2023}\natexlab{}.
\newblock \showarticletitle{We’re Afraid Language Models Aren’t Modeling Ambiguity}. In \bibinfo{booktitle}{\emph{Proc.\ Conference on Empirical Methods in Natural Language Processing (EMNLP)}}. \bibinfo{address}{Singapore}, \bibinfo{pages}{790--807}.
\newblock


\bibitem[Majambere(2011)]%
        {majambere2011clarity}
\bibfield{author}{\bibinfo{person}{Esther Majambere}.} \bibinfo{year}{2011}\natexlab{}.
\newblock \showarticletitle{Clarity, precision and unambiguity: Aspects for effective legislative drafting}.
\newblock \bibinfo{journal}{\emph{Commonwealth Law Bulletin}} \bibinfo{volume}{37}, \bibinfo{number}{3} (\bibinfo{year}{2011}), \bibinfo{pages}{417--426}.
\newblock


\bibitem[Moro and Navigli(2015)]%
        {moro-navigli-2015-semeval}
\bibfield{author}{\bibinfo{person}{Andrea Moro} {and} \bibinfo{person}{Roberto Navigli}.} \bibinfo{year}{2015}\natexlab{}.
\newblock \showarticletitle{{S}em{E}val-2015 Task 13: Multilingual All-Words Sense Disambiguation and Entity Linking}. In \bibinfo{booktitle}{\emph{Proc.\ 9th International Workshop on Semantic Evaluation ({S}em{E}val)}}. \bibinfo{address}{Denver, CO}, \bibinfo{pages}{288--297}.
\newblock


\bibitem[Navigli et~al\mbox{.}(2013)]%
        {navigli-etal-2013-semeval}
\bibfield{author}{\bibinfo{person}{Roberto Navigli}, \bibinfo{person}{David Jurgens}, {and} \bibinfo{person}{Daniele Vannella}.} \bibinfo{year}{2013}\natexlab{}.
\newblock \showarticletitle{{S}em{E}val-2013 Task 12: Multilingual Word Sense Disambiguation}. In \bibinfo{booktitle}{\emph{Proc.\ 7th International Workshop on Semantic Evaluation ({S}em{E}val)}}. \bibinfo{address}{Atlanta, GA}, \bibinfo{pages}{222--231}.
\newblock


\bibitem[Park et~al\mbox{.}(2016)]%
        {park2016multimodal}
\bibfield{author}{\bibinfo{person}{Sunghyun Park}, \bibinfo{person}{Han~Suk Shim}, \bibinfo{person}{Moitreya Chatterjee}, \bibinfo{person}{Kenji Sagae}, {and} \bibinfo{person}{Louis-Philippe Morency}.} \bibinfo{year}{2016}\natexlab{}.
\newblock \showarticletitle{Multimodal analysis and prediction of persuasiveness in online social multimedia}.
\newblock \bibinfo{journal}{\emph{ACM Transactions on Interactive Intelligent Systems}} \bibinfo{volume}{6}, \bibinfo{number}{3}, Article \bibinfo{articleno}{25} (\bibinfo{year}{2016}), \bibinfo{numpages}{25}~pages.
\newblock


\bibitem[Radford et~al\mbox{.}(2023)]%
        {radford2023robust}
\bibfield{author}{\bibinfo{person}{Alec Radford}, \bibinfo{person}{Jong~Wook Kim}, \bibinfo{person}{Tao Xu}, \bibinfo{person}{Greg Brockman}, \bibinfo{person}{Christine McLeavey}, {and} \bibinfo{person}{Ilya Sutskever}.} \bibinfo{year}{2023}\natexlab{}.
\newblock \showarticletitle{Robust speech recognition via large-scale weak supervision}. In \bibinfo{booktitle}{\emph{Proc.\ International Conference on Machine Learning (ICML)}}. \bibinfo{address}{Honolulu, HI}, \bibinfo{pages}{28492--28518}.
\newblock


\bibitem[Rastogi et~al\mbox{.}(2020)]%
        {rastogi2020towards}
\bibfield{author}{\bibinfo{person}{Abhinav Rastogi}, \bibinfo{person}{Xiaoxue Zang}, \bibinfo{person}{Srinivas Sunkara}, \bibinfo{person}{Raghav Gupta}, {and} \bibinfo{person}{Pranav Khaitan}.} \bibinfo{year}{2020}\natexlab{}.
\newblock \showarticletitle{Towards scalable multi-domain conversational agents: The schema-guided dialogue dataset}. In \bibinfo{booktitle}{\emph{Proc.\ the AAAI Conference on Artificial Intelligence (AAAI)}}. \bibinfo{address}{New York, NY}, \bibinfo{pages}{8689--8696}.
\newblock


\bibitem[Ross et~al\mbox{.}(2024)]%
        {ross2024modern}
\bibfield{author}{\bibinfo{person}{Claudia Ross}, \bibinfo{person}{Jing-heng~Sheng Ma}, \bibinfo{person}{Pei-Chia Chen}, \bibinfo{person}{Baozhang He}, {and} \bibinfo{person}{Meng Yeh}.} \bibinfo{year}{2024}\natexlab{}.
\newblock \bibinfo{booktitle}{\emph{Modern Mandarin Chinese Grammar: A Practical Guide}}.
\newblock \bibinfo{publisher}{Routledge}.
\newblock


\bibitem[Schuller et~al\mbox{.}(2017)]%
        {schuller2017interspeech}
\bibfield{author}{\bibinfo{person}{Bj{\"o}rn Schuller}, \bibinfo{person}{Stefan Steidl}, \bibinfo{person}{Anton Batliner}, \bibinfo{person}{Elika Bergelson}, \bibinfo{person}{Jarek Krajewski}, \bibinfo{person}{Christoph Janott}, {et~al\mbox{.}}} \bibinfo{year}{2017}\natexlab{}.
\newblock \showarticletitle{The {INTERSPEECH} 2017 computational paralinguistics challenge: Addressee, cold \& snoring}. In \bibinfo{booktitle}{\emph{Proc.\ the Annual Conference of the International Speech Communication Association (INTERSPEECH)}}. \bibinfo{address}{Stockholm, Sweden}, \bibinfo{pages}{3442--3446}.
\newblock


\bibitem[Szemer{\'e}nyi and Szemer{\'e}nyi(1999)]%
        {szemerenyi1999introduction}
\bibfield{author}{\bibinfo{person}{Oswald Szemer{\'e}nyi} {and} \bibinfo{person}{Oswald John~Louis Szemer{\'e}nyi}.} \bibinfo{year}{1999}\natexlab{}.
\newblock \bibinfo{booktitle}{\emph{Introduction to Indo-European Linguistics}}.
\newblock \bibinfo{publisher}{OUP Oxford}.
\newblock


\bibitem[Tesema et~al\mbox{.}(2023)]%
        {tesema2023addressee}
\bibfield{author}{\bibinfo{person}{Fiseha~B Tesema}, \bibinfo{person}{Jason Gu}, \bibinfo{person}{Wei Song}, \bibinfo{person}{Hong Wu}, \bibinfo{person}{Shiqiang Zhu}, \bibinfo{person}{Zheyuan Lin}, {et~al\mbox{.}}} \bibinfo{year}{2023}\natexlab{}.
\newblock \showarticletitle{Addressee detection using facial and audio features in mixed human--human and human--robot settings: A deep learning framework}.
\newblock \bibinfo{journal}{\emph{IEEE Systems, Man, and Cybernetics Magazine}} \bibinfo{volume}{9}, \bibinfo{number}{2} (\bibinfo{year}{2023}), \bibinfo{pages}{25--38}.
\newblock


\bibitem[Wang et~al\mbox{.}(2024a)]%
        {wang-etal-2024-disambiguate}
\bibfield{author}{\bibinfo{person}{Yue Wang}, \bibinfo{person}{Qiliang Liang}, \bibinfo{person}{Yaqi Yin}, \bibinfo{person}{Hansi Wang}, {and} \bibinfo{person}{Yang Liu}.} \bibinfo{year}{2024}\natexlab{a}.
\newblock \showarticletitle{Disambiguate words like composing them: A morphology-informed approach to enhance {C}hinese word sense disambiguation}. In \bibinfo{booktitle}{\emph{Proc.\ 62nd Annual Meeting of the Association for Computational Linguistics (ACL)}}. \bibinfo{address}{Bangkok, Thailand}, \bibinfo{pages}{15354--15365}.
\newblock


\bibitem[Wang et~al\mbox{.}(2024b)]%
        {wang2024morpheme}
\bibfield{author}{\bibinfo{person}{Yue Wang}, \bibinfo{person}{Hua Zheng}, \bibinfo{person}{Yaqi Yin}, \bibinfo{person}{Hansi Wang}, \bibinfo{person}{Qiliang Liang}, {and} \bibinfo{person}{Yang Liu}.} \bibinfo{year}{2024}\natexlab{b}.
\newblock \showarticletitle{Morpheme Sense Disambiguation: A New Task Aiming for Understanding the Language at Character Level}. In \bibinfo{booktitle}{\emph{Proceedings of the Joint International Conference on Computational Linguistics, Language Resources and Evaluation (LREC-COLING)}}. \bibinfo{address}{Turin, Italy}, \bibinfo{pages}{11605--11618}.
\newblock


\bibitem[Wu et~al\mbox{.}(2021)]%
        {wu2021mumor}
\bibfield{author}{\bibinfo{person}{Jiaming Wu}, \bibinfo{person}{Hongfei Lin}, \bibinfo{person}{Liang Yang}, {and} \bibinfo{person}{Bo Xu}.} \bibinfo{year}{2021}\natexlab{}.
\newblock \showarticletitle{Mumor: A multimodal dataset for humor detection in conversations}. In \bibinfo{booktitle}{\emph{Proc.\ 10th CCF International Conference on Natural Language Processing and Chinese Computing (NLPCC)}}. \bibinfo{address}{Qingdao, China}, \bibinfo{pages}{619--627}.
\newblock


\bibitem[Xu et~al\mbox{.}(2025)]%
        {xu2025qwen2}
\bibfield{author}{\bibinfo{person}{Jin Xu}, \bibinfo{person}{Zhifang Guo}, \bibinfo{person}{Jinzheng He}, \bibinfo{person}{Hangrui Hu}, \bibinfo{person}{Ting He}, \bibinfo{person}{Shuai Bai}, {et~al\mbox{.}}} \bibinfo{year}{2025}\natexlab{}.
\newblock \showarticletitle{Qwen2. 5-omni technical report}.
\newblock \bibinfo{journal}{\emph{arXiv preprint arXiv:2503.20215}} (\bibinfo{year}{2025}).
\newblock


\bibitem[Yan et~al\mbox{.}(2023)]%
        {yan-etal-2023-ji}
\bibfield{author}{\bibinfo{person}{Fukang Yan}, \bibinfo{person}{Yue Zhang}, {and} \bibinfo{person}{Zhenghua Li}.} \bibinfo{year}{2023}\natexlab{}.
\newblock \showarticletitle{Construction of a modern {C}hinese word sense dataset based on online dictionaries}. In \bibinfo{booktitle}{\emph{Proc.\ 22nd Chinese National Conference on Computational Linguistics}}. \bibinfo{address}{Harbin, China}, \bibinfo{pages}{43--53}.
\newblock


\bibitem[Yue et~al\mbox{.}(2024)]%
        {yue-etal-2024-sarcnet}
\bibfield{author}{\bibinfo{person}{Tan Yue}, \bibinfo{person}{Xuzhao Shi}, \bibinfo{person}{Rui Mao}, \bibinfo{person}{Zonghai Hu}, {and} \bibinfo{person}{Erik Cambria}.} \bibinfo{year}{2024}\natexlab{}.
\newblock \showarticletitle{{S}arc{N}et: A multilingual multimodal sarcasm detection dataset}. In \bibinfo{booktitle}{\emph{Proc.\ Joint International Conference on Computational Linguistics, Language Resources and Evaluation (LREC-COLING)}}. \bibinfo{address}{Turin, Italy}, \bibinfo{pages}{14325--14335}.
\newblock


\bibitem[Zhang et~al\mbox{.}(2024a)]%
        {zhang2024chambi}
\bibfield{author}{\bibinfo{person}{Qin Zhang}, \bibinfo{person}{Sihan Cai}, \bibinfo{person}{Jiaxu Zhao}, \bibinfo{person}{Mykola Pechenizkiy}, {and} \bibinfo{person}{Meng Fang}.} \bibinfo{year}{2024}\natexlab{a}.
\newblock \showarticletitle{CHAmbi: A new benchmark on {C}hinese ambiguity challenges for large language models}. In \bibinfo{booktitle}{\emph{Proc.\ Findings of the Association for Computational Linguistics: EMNLP}}. \bibinfo{address}{Miami, FL}, \bibinfo{pages}{14883--14898}.
\newblock


\bibitem[Zhang et~al\mbox{.}(2024b)]%
        {zhang2024refashioning}
\bibfield{author}{\bibinfo{person}{Zixing Zhang}, \bibinfo{person}{Liyizhe Peng}, \bibinfo{person}{Tao Pang}, \bibinfo{person}{Jing Han}, \bibinfo{person}{Huan Zhao}, {and} \bibinfo{person}{Bj{\"o}rn~W Schuller}.} \bibinfo{year}{2024}\natexlab{b}.
\newblock \showarticletitle{Refashioning emotion recognition modelling: The advent of generalised large models}.
\newblock \bibinfo{journal}{\emph{IEEE Transactions on Computational Social Systems}} \bibinfo{volume}{11}, \bibinfo{number}{5} (\bibinfo{year}{2024}), \bibinfo{pages}{6690--6704}.
\newblock


\bibitem[Zhang et~al\mbox{.}(2024c)]%
        {zhang2024paralbench}
\bibfield{author}{\bibinfo{person}{Zixing Zhang}, \bibinfo{person}{Weixiang Xu}, \bibinfo{person}{Zhongren Dong}, \bibinfo{person}{Kanglin Wang}, \bibinfo{person}{Yimeng Wu}, \bibinfo{person}{Jing Peng}, {et~al\mbox{.}}} \bibinfo{year}{2024}\natexlab{c}.
\newblock \showarticletitle{ParaLBench: A large-scale benchmark for computational paralinguistics over acoustic foundation models}.
\newblock \bibinfo{journal}{\emph{IEEE Transactions on Affective Computing}} (\bibinfo{year}{2024}).
\newblock
\newblock
\shownote{In press, 17 pages}.


\bibitem[Zheng et~al\mbox{.}(2021)]%
        {zheng2021leveraging}
\bibfield{author}{\bibinfo{person}{Hua Zheng}, \bibinfo{person}{Lei Li}, \bibinfo{person}{Damai Dai}, \bibinfo{person}{Deli Chen}, \bibinfo{person}{Tianyu Liu}, \bibinfo{person}{Xu Sun}, {and} \bibinfo{person}{Yang Liu}.} \bibinfo{year}{2021}\natexlab{}.
\newblock \showarticletitle{Leveraging word-formation knowledge for Chinese word sense disambiguation}. In \bibinfo{booktitle}{\emph{Proc.\ Findings of the Association for Computational Linguistics: EMNLP}}. \bibinfo{address}{Punta Cana, Dominican Republic}, \bibinfo{pages}{918--923}.
\newblock


\end{thebibliography}
